%% file: main.tex
\documentclass{article}

\usepackage{microtype}
\usepackage{graphicx}
\usepackage{subcaption}  % replaces deprecated subfigure
\usepackage{booktabs}
\usepackage{amsmath,amssymb,amsfonts}
\usepackage{algorithmic}
\newcommand{\RETURN}{\textbf{return}~}
\usepackage{textcomp}
\usepackage{xcolor}
\usepackage{makecell}
\usepackage{multirow}
\usepackage{diagbox}
\usepackage{balance}
\usepackage[flushleft]{threeparttable}
\usepackage[T1]{fontenc}    
\usepackage{hyperref}
\renewcommand{\algorithmiccomment}[1]{\hfill $\triangleright$ \textit{#1}}
\usepackage{url}
\usepackage{nicefrac}
\usepackage{listings}
\usepackage{adjustbox}
\usepackage{algorithm}
\usepackage{enumitem}

\usepackage{pifont}

\newtheorem{theorem}{Theorem}

\newtheorem{assumption}{Assumption}

\newcommand{\toolname}[0]{\textsc{KernelBand}}

\usepackage[preprint]{icml2026}

\icmltitlerunning{Steering LLM-based Kernel Optimization with \toolname{}}

\begin{document}

\twocolumn[
  \icmltitle{\toolname{}:Steering LLM-based Kernel Optimization via Hardware-Aware Multi-Armed Bandits}

  \icmlsetsymbol{equal}{*}

  \begin{icmlauthorlist}
    \icmlauthor{Dezhi Ran}{equal,scs,tongming}
    \icmlauthor{Shuxiao Xie}{equal,tongming,ecnu}
    \icmlauthor{Mingfang Ji}{tongming,tju}
    \icmlauthor{Anmin Liu}{scs}
    \icmlauthor{Mengzhou Wu}{scs,tongming}
    \icmlauthor{Yuan Cao}{scs,tongming}
    \icmlauthor{Yuzhe Guo}{scs,tongming}
    \icmlauthor{Hao Yu}{hkust}
    \icmlauthor{Linyi Li}{sfu}
    \icmlauthor{Yitao Hu}{tju}
    \icmlauthor{Wei Yang}{dallas}
    \icmlauthor{Tao Xie}{scs,tongming,fisac,sisoc}
  \end{icmlauthorlist}

  \icmlaffiliation{scs}{Key Lab of HCST (PKU), MOE; SCS, Peking University, Beijing, China}
  \icmlaffiliation{hkust}{Hong Kong University of Science and Technology, Hong Kong, China}
  \icmlaffiliation{ecnu}{East China Normal University, Shanghai, China}
  \icmlaffiliation{tju}{Department of Computer Science, Tianjin University, Tianjin, China}
  \icmlaffiliation{sfu}{School of Computing Science, Simon Fraser University, Burnaby, BC, Canada}
  \icmlaffiliation{dallas}{University of Texas at Dallas, USA}
  \icmlaffiliation{tongming}{Beijing Tongming Lake Information Technology Application Innovation Center, China}
  \icmlaffiliation{fisac}{Fudan University Institute of Systems for Advanced Computing, China}
  \icmlaffiliation{sisoc}{Shanghai Institute of Systems for Open Computing, China}
  \icmlcorrespondingauthor{Tao Xie}{taoxie@pku.edu.cn}

  \icmlkeywords{Code Generation, Large Language Models, Benchmark}

  \vskip 0.3in
]
\printAffiliationsAndNotice{\icmlEqualContribution}

\begin{abstract}
High-performance GPU kernels are critical for efficient LLM serving, yet their optimization remains a bottleneck requiring deep system expertise. While code LLMs show promise in generating functionally correct code, kernel optimization is intrinsically a search problem over a vast optimization space. 
The fundamental mismatch prevents existing LLM agents from efficiently exploring the optimization space for diverse hardware and compute patterns. 
To bridge the gap, we present \toolname{}, a framework that formulates kernel optimization as a Multi-Armed Bandit (MAB) problem, explicitly balancing exploration and exploitation to unlock the potential of code LLMs. 
To navigate the infinite arm space of optimization strategies applied to candidate kernels, we design two key mechanisms: a hardware-aware pruning strategy via profiling bounds and a trace-driven clustering algorithm that leverages Lipschitz continuity.
Theoretically, we prove that \toolname{} reduces the regret bound to depend on the compact covering number of runtime clusters, ensuring sample-efficient discovery of high-performance kernels.
Extensive experiments on TritonBench-G with three GPU architectures and four code LLMs show that \toolname{} consistently and substantially outperforms state-of-the-art methods with over 33\% average improvement.
\end{abstract}

\input{introduction}
\input{problem}

\input{method}

\input{experiment}

\input{related_work}
\input{conclusion}

\bibliography{main.bib}
\bibliographystyle{icml2026}
\input{appendix}
\balance
\end{document}

%% file: introduction.tex
\section{Introduction}
The computational demands of Large Language Models (LLMs) have grown exponentially~\citep{zhao2023survey,naveed2025comprehensive,floridi2020gpt,team2025kimi,team2023gemini,bai2023qwen}, making efficient serving infrastructure a critical priority~\citep{miao2025towards,ye2025flashinfer,kwon2023efficient,fang2021turbotransformers,yan2018efficient,lin2024infinite,pan2024instinfer}. At the heart of the efficiency lies kernel optimization~\citep{filipovivc2015optimizing,ryoo2008optimization,lange2025towards}, the engineering of high-performance primitives for fundamental operations such as attention~\citep{dong2024flex} and GEMM~\citep{faingnaert2021flexible}. 
Traditionally, developing these kernels has been the exclusive domain of specialized experts, requiring careful manual mapping of algorithms to complex hardware features, such as multi-level memory hierarchies and tensor core instructions~\citep{nvid25b}. 
To democratize this process, domain-specific languages like Triton~\citep{tillet2019triton} and TileLang~\citep{wang2025tilelang} have emerged, offering abstractions that hide low-level intricacies. However, while DSLs simplify implementation, achieving peak performance still relies heavily on compiler autotuning mechanisms to select configuration parameters~\citep{gao2025automatic,zhu2022roller,wang2024ladder}. 
As hardware complexity increases, the configuration space suffers from a combinatorial explosion~\citep{gao2025automatic,ansel2024pytorch,shi2023welder}, rendering heuristic strategies computationally prohibitive and often incapable of finding global optima within reasonable timeframes.

To circumvent the preceding limitation, the community has recently turned to code LLMs~\citep{dong2025survey,guo2024deepseek,hui2024qwen2,achiam2023gpt,caruccio2024claude}, leveraging their generative capabilities to automate kernel optimization. 
Current methods primarily fall into two categories: agent-based methods~\citep{dong2025starks,wang2025geak,cudaforge}, which use iterative feedback loops to refine implementations, and training-based methods~\citep{baronio2025kevin,woo2025tritonrl,kong2025concur}, which fine-tune models on optimization-specific datasets. 
However, a fundamental mismatch persists: LLMs are inherently trained to generate statistically probable and functionally correct code, whereas kernel optimization is intrinsically a search problem over a vast and discontinuous optimization space~\citep{gao2025automatic}. Consequently, existing agents struggle to balance exploration and exploitation, frequently converging to suboptimal local minima or wasting computational resources on invalid configurations, thereby failing to achieve expert-level performance.

To bridge the gap, we present \toolname{}, the first Multi-Armed Bandit (MAB)~\citep{mahajan2008multi,scott2010modern,boursier2024survey,silva2022multi} framework to guide LLMs in navigating the optimization space with provable efficiency, enabled by two novel designs.
Distinguished from previous methods~\citep{cudaforge,kong2025concur} that rely on unguided generation or self-reflection heuristics, \toolname{} formulates the problem in a contextual bandit setting~\cite{bouneffouf2020survey}, where each arm corresponds to the application of an optimization strategy~\citep{ryoo2008optimization} (e.g., tiling, vectorization) to a candidate kernel implementation.
To address the challenge of infinite action spaces, we design hardware-aware pruning, which utilizes profiling data to establish tight reward upper bounds, and trace-driven clustering, which exploits the Lipschitz continuity~\citep{hager1979lipschitz} of runtime behaviors to estimate rewards for unexplored arms.
Theoretically, we prove that \toolname{} reduces the regret bound to depend on the compact covering number of runtime clusters rather than the vast kernel space.

We evaluate \toolname{} on TritonBench-G~\citep{tritonbench} across three GPU architectures (RTX 4090, H20, A100) and four frontier code LLMs~\citep{deepseek_v3_2, openai_gpt5, anthropic_claude_opus45, google_gemini3_flash}. Empirical results demonstrate consistent superiority over state-of-the-art baselines, achieving up to \textbf{1.91$\times$} geometric mean speedup and improving the Fast@1 success rate by \textbf{39--140\%}. Ablation studies confirm that structured exploration is foundational: replacing our bandit policy with LLM semantic reasoning regresses performance to 0.97$\times$ (below the reference kernel), validating that learned execution statistics outperform intuition. Furthermore, \toolname{} automatically adapts strategies to hardware bottlenecks and delivers \textbf{35--50\%} higher speedup per dollar than unguided approaches.

This paper makes the following main contributions:
\begin{itemize}
    \item We propose \toolname{}, the first framework to formulate LLM-based kernel optimization as a MAB problem, effectively resolving the mismatch between functional code generation and optimization search.
    \item We design a hardware-aware acquisition strategy that combines profiling-based pruning with trace-driven clustering, proved to efficiently explore the kernel optimization space.
    \item Extensive experiments on TritonBench-G across three GPUs and four code LLMs demonstrate that \toolname{} consistently outperforms state-of-the-art methods, achieving up to \textbf{1.91$\times$} geometric mean speedup, with ablation results confirming the structured bandit policy as the key to the performance.
\end{itemize}

%% file: problem.tex
\section{Problem Formulation}
\label{sec:formulation}

In this section, we formalize kernel optimization as a search problem over a generated code space and subsequently model it as a structured contextual MAB problem with an expanding action space.
We explicitly address the  mismatch between the generative capabilities of code LLMs (which prioritize functional correctness) and the navigational requirements of kernel optimization (which is intrinsically a search problem over performance landscapes).

\subsection{The Search Problem}
\label{subsec:search-problem}
Given a target hardware platform $\mathcal{H}$ and a naive kernel implementation, we seek a kernel $k^*$ minimizing latency $L(k,\mathcal{H})$ while preserving functional correctness:
\begin{equation}
    k^* = \operatorname*{argmin}_{k \in \mathcal{K}_{\text{valid}}} L(k, \mathcal{H})
\end{equation}
where $\mathcal{K}_{\text{valid}}$ contains all correct implementations. 

We view optimization as traversing a directed graph $\mathcal{G}=(\mathcal{V},\mathcal{E})$: nodes $\mathcal{V}$ are valid kernels; edges $(k\to k')\in\mathcal{E}$ represent applying a strategy $s\in\mathcal{S}$ via a code LLM.

\begin{figure*}[t]
    \centering
    \includegraphics[width=.85\linewidth]{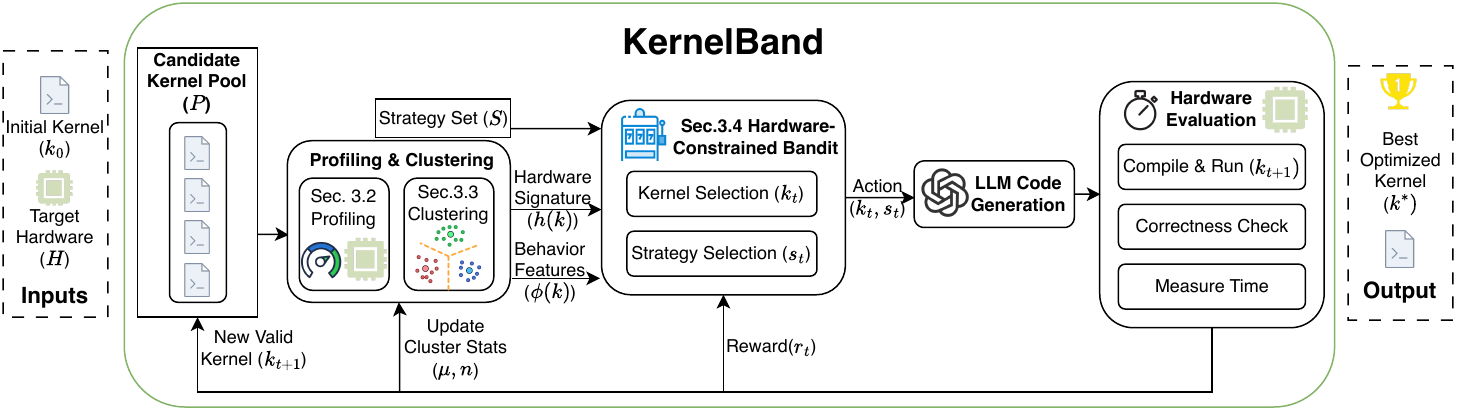}
    \caption{Overview of \toolname{}. \toolname{} interleaves runtime analysis (clustering) with hardware-aware decision making (masked bandit) to guide LLM-based kernel optimization.}
    \label{fig:overview}
\end{figure*}

\paragraph{The Fundamental Mismatch.} While modern Code LLMs excel at generating functionally correct nodes (kernels), they lack the hardware-specific intuition needed to select edges that efficiently navigate toward performance-optimal regions of $\mathcal{G}$. A naive LLM-based optimizer performs what amounts to a random walk on the graph, wasting substantial efforts on transformations that yield negligible or negative speedups. Our goal is to replace this undirected exploration with a principled decision policy that leverages both program semantics and hardware behavior.

\subsection{Contextual Bandit Formulation with Expanding Action Space}
\label{subsec:bandit-formulation}

We frame kernel optimization as an \textbf{iterative frontier expansion} process. Starting from an initial implementation $\mathcal{P}_0 = \{k_{\text{naive}}\}$, at each step $t$, the agent maintains a frontier $\mathcal{P}_t \subseteq \mathcal{V}$ of promising kernels discovered so far. The core decision is: \textit{which kernel $k \in \mathcal{P}_t$ to expand using which optimization strategy $s \in \mathcal{S}$?}

We formulate this as a \textbf{contextual bandit problem} rather than full Reinforcement Learning (RL) for three key reasons: first, optimization trajectories are short, making immediate reward predictive of final performance; second, LLM generation variance complicates long-horizon value estimation; and third, bandits offer rigorous regret bounds under our structural assumptions, ensuring sample efficiency.

\paragraph{Formal Model.}
We formulate the optimization process as a contextual bandit defined by a tuple of context, action, transition, and reward.
The \textbf{context space} consists of vectors $x_k \in \mathbb{R}^d$ for each candidate $k \in \mathcal{P}_t$, encoding both static program features and dynamic execution traces $\phi(k)$.
The \textbf{action space} at time $t$, denoted $\mathcal{A}_t = \mathcal{P}_t \times \mathcal{S}$, represents applying strategy $s$ to kernel $k$. Crucially, since the frontier $|\mathcal{P}_t|$ grows with $t$, $\mathcal{A}_t$ is unbounded, distinguishing this from standard fixed-arm bandits.
Executing action $a_t = (k_t, s_t)$ triggers a \textbf{generative transition} $k'_t \sim P_{\text{LLM}}(\cdot \mid k_t, s_t, \mathcal{H})$, where stochasticity stems from the LLM's sampling process.
Finally, the agent observes a \textbf{reward signal} $r_t \in [0,1]$ measuring normalized latency improvement: $r_t = \text{Clip}(\frac{L(k_t) - L(k'_t)}{L(k_t)}, 0, 1)$, where zero reward is assigned to performance regressions or compilation failures.
The agent's objective is to maximize the cumulative expected reward $\sum_{t=1}^T \mathbb{E}[r_t]$.

\subsection{Structural Properties Enabling Tractable Learning}
\label{subsec:structure}

The expanding action space $\mathcal{A}_t$ poses a fundamental challenge: standard bandit algorithms would suffer linear regret $O(T)$. However, kernel optimization exhibits two key structural properties that make the problem tractable:

\begin{assumption}[Hardware-Aware Gain Boundedness]
\label{asm:boundedness}
The expected improvement from applying strategy $s$ to kernel $k$ is bounded by hardware limits. Formally, there exists a known bounding function~\citep{williams2009roofline} $B: \mathcal{V} \times \mathcal{S} \to [0,1]$ such that:
\begin{equation}
    \mu(k,s) \triangleq \mathbb{E}[r_t | k, s] \leq B(k,s)
\end{equation}
where $B(k,s)$ estimates the maximum achievable gain based on the gap between $k$'s current performance and the theoretical limit for strategy $s$ on hardware $\mathcal{H}$.
\end{assumption}

\begin{assumption}[Lipschitz Continuity in Behavior Space]
\label{asm:smoothness}
The expected reward function $\mu(k,s)$ is Lipschitz continuous with respect to execution behavior. That is, for any strategy $s$ and kernels $k_1, k_2$:
\begin{equation}
    |\mu(k_1, s) - \mu(k_2, s)| \leq L \cdot \|\phi(k_1) - \phi(k_2)\|_2
\end{equation}
where $\phi(k)$ captures runtime characteristics. This implies that kernels with similar bottlenecks respond similarly to optimizations.
\end{assumption}

\textbf{Intuition.} Together, these assumptions imply that while the code space is infinite, the space of \textit{meaningful optimization decisions} is low-dimensional and structured, motivating the design of \toolname{}.

%% file: method.tex
\section{Methodology: \toolname{}}
\label{sec:approach}

\subsection{System Overview}
\label{subsec:overview}
As shown in Figure~\ref{fig:overview}, \toolname{} addresses the challenge of the expanding action space $\mathcal{A}_t = \mathcal{P}_t \times \mathcal{S}$ through three key components: (1) \textbf{Runtime behavior characterization} (Sec~\ref{subsec:characterization}), which extracts execution signatures to enable knowledge sharing; (2) \textbf{Dynamic clustering} (Sec~\ref{subsec:clustering}), which groups similar kernels to manage the expanding action space; and (3) \textbf{Hardware-constrained bandit policy} (Sec~\ref{subsec:bandit}), which selects promising optimizations while pruning physically invalid strategies.

\subsection{Runtime Behavior Characterization}
\label{subsec:characterization}

We characterize each kernel $k$ using two complementary representations to satisfy the smoothness and boundedness assumptions defined in Section~\ref{sec:formulation}.
\paragraph{Behavioral feature vector $\phi(k)$.}
For clustering kernels with similar optimization responses, we define a 5-dimensional vector $\phi(k)$ comprising normalized execution time $\tilde{\mathcal{T}}(k)$ and hardware counters:
\begin{equation}
    \phi(k) = [\tilde{\mathcal{T}}(k),\ n_{\text{reg}},\ n_{\text{smem}},\ d_{\text{block}},\ \eta_{\text{occ}}]
\end{equation}
representing registers per thread, shared memory per block, block dimension, and occupancy, respectively. Kernels close in $\phi$-space share similar bottlenecks (Assumption~\ref{asm:smoothness}), allowing the bandit to generalize strategy performance.

\paragraph{Hardware signature $h(k)$.}
To implement the bounding function $B(k,s)$ from Assumption~\ref{asm:boundedness}, we extract a hardware signature $h(k)$ using NVIDIA Nsight Compute~\citep{nsight_compute}, measuring peak throughput percentages for DRAM, L2 cache, and SM. These metrics identify the dominant bottleneck (memory bandwidth/compute/cache) for pruning physically implausible optimization strategies.

\subsection{Structured Exploration via Dynamic Clustering}
\label{subsec:clustering}

\paragraph{Periodic re-clustering.}
Instead of maintaining a separate bandit arm for every kernel, we maintain arms for kernel \textit{clusters}. At iteration $t$, we partition the frontier $\mathcal{P}_t$ into $K$ clusters $\mathcal{C}_t = \{C_1, \dots, C_K\}$ using K-Means on $\{\phi(k)\}$. 
Clusters are recomputed every $\tau$ iterations. This periodic update balances the need to track shifting kernel behaviors with the stability required for bandit convergence.

\paragraph{Representative profiling.}
Since hardware profiling is expensive ($\approx 10$s), we profile only the centroid kernel $k_c^{(i)}$ of each active cluster during the re-clustering phase. We approximate the hardware constraints of the entire cluster using this representative, significantly reducing overhead.

\subsection{Hardware-Constrained Bandit Policy}
\label{subsec:bandit}

As depicted in Algorithm~\ref{alg:main}, we formulate the decision process as a \textit{masked bandit}.

\paragraph{Pruning via hardware potential.}
We define a binary mask $M_{i,s} \in \{0,1\}$ based on the hardware signature $h(k)$. A strategy $s$ is valid for cluster $C_i$ only if it targets a non-saturated resource:
\begin{equation}
    M_{i,s} = \mathbb{I}\left[ h(k_c^{(i)})[\text{Target}(s)] < \theta_{\text{sat}} \right]
\end{equation}
where $\theta_{\text{sat}}$ is a saturation threshold (e.g., 75\%). This reduces the effective action space from $|\mathcal{S}|$ to $|\mathcal{S}_{\text{valid}}|$.

\paragraph{Action selection.}
At iteration $t$, we select a cluster-strategy pair $(I_t, S_t)$ using a Masked UCB index. Let $\hat{\mu}_{i,s}$ be the empirical mean reward and $N_{i,s}$ the visit count. We maximize UCB only among valid actions:
\begin{equation}
\label{eq:masked_ucb}
    (I_t, S_t) = \operatorname*{argmax}_{\substack{i \in [K], s \in \mathcal{S} \\ \text{s.t. } M_{i,s}=1}} \left( \hat{\mu}_{i,s} + c \sqrt{\frac{\ln t}{N_{i,s}}} \right)
\end{equation}
Once $(I_t, S_t)$ is chosen, we sample a specific kernel $k_t \in C_{I_t}$ via a softmax distribution over local potential scores $V_{\text{hw}}(k, S_t) = \theta_{\text{sat}} - h(k)[\text{Target}(S_t)]$, measuring the remaining optimization headroom for strategy $S_t$ on kernel $k$.

\begin{algorithm}[t]
\caption{Workflow of \toolname{}}
\label{alg:main}
\begin{algorithmic}[1]
\REQUIRE Kernel $k_0$, Strategies $\mathcal{S}$, Budget $T$, Period $\tau=10$, Clusters $K$
\STATE $\mathcal{P} \leftarrow \{k_0\}$, $\mathcal{C} \leftarrow \{\mathcal{P}\}$ \COMMENT{Initial single cluster}
\STATE Initialize $N_{i,s} \leftarrow 1, \hat{\mu}_{i,s} \leftarrow 0.5$ for all $i,s$
\STATE Initialize Masks $M_{i,s} \leftarrow 1$ \COMMENT{No pruning initially}

\FOR{$t = 1$ \textbf{to} $T$}
    \STATE Compute $\phi(k)$ for all $k \in \mathcal{P}$
    \STATE \COMMENT{\textbf{Periodic clustering \& profiling}}
    \IF{$t \bmod \tau = 0$ \AND $|\mathcal{P}| \geq 2K$} 
        \STATE $\mathcal{C} \leftarrow \text{KMeans}(\{\phi(k)\}, K)$
        \STATE Update centroids $k_c^{(i)}$ and profile $h_c^{(i)} \leftarrow \text{NCU}(k_c^{(i)})$
    \ENDIF
    
    \STATE \COMMENT{\textbf{Hardware-constrained selection}}
    \IF{centroids are profiled}
        \STATE $M_{i,s} \leftarrow \mathbb{I}[h_c^{(i)}[\text{Target}(s)] < \theta_{\text{sat}}]$
    \ENDIF
    \STATE Select $(I_t, S_t)$ via Eq.~\eqref{eq:masked_ucb} (Masked UCB)
    \STATE Sample $k_t \in C_{I_t}$ with $P(k) \propto \exp(V_{\text{hw}}(k, S_t))$
    
    \STATE \COMMENT{\textbf{Generation \& update}} 
    \STATE $k'_t \leftarrow \text{LLM}(k_t, S_t)$
    \IF{$\text{Verify}(k'_t)$}
        \STATE $r_t \leftarrow \max(0, (\mathcal{T}(k_t) - \mathcal{T}(k'_t))/\mathcal{T}(k_t))$
        \STATE $\mathcal{P} \leftarrow \mathcal{P} \cup \{k'_t\}$
        \STATE $N_{I_{t}, S_{t}} \leftarrow N_{I_{t}, S_{t}} + 1$
        \STATE $\hat{\mu}_{I_{t}, S_{t}} \leftarrow \hat{\mu}_{I_{t}, S_{t}} + \frac{r_t - \hat{\mu}_{I_{t}, S_{t}}}{N_{I_{t}, S_{t}}}$
    \ENDIF
\ENDFOR

\RETURN $\arg\min_{k \in \mathcal{P}} \mathcal{T}(k)$
\end{algorithmic}
\end{algorithm}

\subsection{Theoretical Analysis}
\label{subsec:theory}

We analyze \toolname{} through the lens of \textbf{$\epsilon$-optimality}, acknowledging that in high-dimensional kernel optimization, finding a solution within a small tolerance of the global optimum is the practical goal.

\begin{theorem}[Convergence to $\epsilon$-Optimal Solution]
\label{thm:convergence}
Let $\mu^*$ be the optimal performance and $r_t$ be the reward at step $t$. Under Assumptions~\ref{asm:boundedness} and~\ref{asm:smoothness}, with probability $1-\delta$, the average regret satisfies:
\begin{equation}
\frac{1}{T}\sum_{t=1}^T \mathbb{E}[\mu^* - r_t] \ \leq \ C\sqrt{\frac{K|\mathcal{S}_{\text{valid}}|\ln T}{T}} \ + \ L \cdot \max_i \text{diam}(C_i)
\end{equation}
where $C$ is a constant derived from the UCB analysis.
\end{theorem}

Due to space limit, we put the proof in Appendix~\ref{app:theory}.

\subsection{Implementation Details}
\label{subsec:implementation}

We employ $|\mathcal{S}|=6$ optimization strategies: \textit{tiling}, \textit{vectorization}, \textit{fusion}, \textit{pipeline}, \textit{reordering}, and \textit{access \& layout} (details in Appendix~\ref{app:strategies}). Kernel clustering uses scikit-learn's KMeans with $K=3$ clusters, a value empirically validated in our experiments; clusters are recomputed every $\tau=10$ iterations to balance adaptability with computational overhead. For hardware-informed pruning, we use NVIDIA Nsight Compute (NCU) to record throughput metrics (caching results by code hash), and apply a saturation threshold $\theta_{\text{sat}}=75\%$ to filter strategies that target already-saturated resources. The UCB exploration parameter is $c=2.0$. These design choices introduce minimal overhead: feature extraction adds less than 1\% runtime cost, and clustering incurs negligible cost, scaling as $O(|\mathcal{P}_t|)$ every $\tau$ iterations. The dominant costs remain LLM generation and kernel compilation (Section~\ref{sec:cost}); parameter sensitivity is analyzed in Sections~\ref{sec:scaling} and~\ref{sec:ablation}.

%% file: experiment.tex
\section{Experiments}
\subsection{Experimental Settings}

\textbf{Benchmark.}
We evaluate on a corrected version of TritonBench-G~\citep{tritonbench}, a popular benchmark of GPU Triton kernels (details in Appendix~\ref{app:baseline_discussion}).
After excluding \texttt{sin\_computation} (which admits trivial simplification yielding artificially high speedups), we obtain 183 kernels spanning 13 functional categories (e.g., attention, matrix multiplication, normalization, fused operations) and 5 difficulty levels (L1-L5).

\textbf{Hardware platforms.}
We conduct experiments on three NVIDIA GPUs covering consumer-grade (RTX 4090) and datacenter-grade (H20 and A100) hardware. All platforms run CUDA 12.1 with Triton 3.3.0.

\textbf{LLM backend.}
We use DeepSeek-V3.2~\citep{deepseek_v3_2} as the primary LLM backend. Detailed model configurations are provided in Appendix~\ref{app:llm_config}. We evaluate model generalization with additional backends~\citep{openai_gpt5, anthropic_claude_opus45, google_gemini3_flash} in Section~\ref{sec:model_gen}.

\textbf{Baselines.}
We use: (1) \textbf{GEAK}~\citep{wang2025geak}, an open-source Triton kernel optimization agent using iterative refinement, with minimal adaptation for our hardware and (2) \textbf{Best-of-N (BoN)}, which samples $N=T$ independent variants and selects the fastest (isolating iterative effects). 
Adaptation details are in Appendix~\ref{app:baseline_discussion}.
We also compare against \textbf{PyTorch} baselines, eager execution, \texttt{torch.compile} with inductor backend, and \texttt{torch.compile} with max-autotune, to contextualize optimization gains with standard PyTorch execution (Appendix~\ref{app:pytorch_baseline}).

\textbf{Optimization budget.}
We optimize each kernel for $T=20$ iterations. Main results (Section~\ref{sec:main_results}) use this default budget; scaling analysis (Section~\ref{sec:scaling}) extends to $T=40$ iterations to study convergence behavior.

\textbf{Evaluation methodology.}
We adopt TritonBench's hierarchical evaluation structure. Each task optimizes a single reference kernel over $T$ iterations. Candidates undergo \textbf{two-stage correctness verification}: \textit{Call Accuracy} checks for runtime errors, while \textit{Execution Accuracy} verifies numerical equivalence via \texttt{torch.allclose}. Passing candidates are benchmarked across 10+ input shapes; per-task speedup is calculated as the \textbf{ratio of total runtimes} (baseline over optimized) for the best correct candidate, naturally prioritizing computationally dominant shapes. Details are provided in Appendix~\ref{app:eval_details}.

\textbf{Metrics.}
We report three complementary metrics:
(1) \textbf{Correct (\%)}: Percentage of tasks yielding at least one valid kernel ($\ge 1$).
(2) \textbf{Fast@1 (\%)}: Percentage of tasks where the best kernel achieves speedup $>1.0\times$ (failed tasks count as 0).
(3) \textbf{Geometric Mean Speedup}: Reported in two modes. \textit{Standard mode} (tables) averages only correct tasks to isolate optimization quality. \textit{Fallback mode} (figures) assigns failures and regressions a baseline speedup of $1.0\times$, ensuring monotonic scaling curves and reflecting practical deployment where users fall back to the reference kernel.

\subsection{Main Results}
\label{sec:main_results}

\begin{table*}[t]
\centering
\caption{Performance on TritonBench-G, stratified by difficulty level (L1:easiest, L5:hardest). Adjacent levels are merged where sample sizes are small (L1: 3, L5: 5 kernels) to ensure statistically meaningful aggregation. C: Correct (\%). F: Fast@1 (\%). G: Geometric mean speedup (\emph{standard mode}: computed over correct tasks only, including regressions). Best results per configuration in \textbf{bold}.}
\label{tab:main_results}
\small
\setlength{\tabcolsep}{6pt}
\begin{tabular}{ll|ccc|ccc|ccc|ccc}
\toprule
& & \multicolumn{3}{c|}{\textbf{L1--2}} & \multicolumn{3}{c|}{\textbf{L3}} & \multicolumn{3}{c|}{\textbf{L4--5}} & \multicolumn{3}{c}{\textbf{All}} \\
\textbf{Platform} & \textbf{Method} & C & F & G & C & F & G & C & F & G & C & F & G \\
\midrule
\multirow{3}{*}{RTX 4090}
& BoN         & 63.6 & 9.1 & 0.86 & 12.1 & 6.1 & 0.89 & 28.6 & 14.3 & 1.09 & 31.1 & 10.0 & 0.96 \\
& GEAK        & 68.2 & 36.4 & 1.62 & 21.2 & 18.2 & 1.45 & 80.0 & 40.0 & 1.34 & 55.6 & 31.1 & 1.44 \\
& \toolname{} & \textbf{81.8} & \textbf{50.0} & \textbf{2.14} & \textbf{60.6} & \textbf{30.3} & \textbf{1.82} & \textbf{91.4} & \textbf{51.4} & \textbf{1.47} & \textbf{77.8} & \textbf{43.3} & \textbf{1.74} \\
\midrule
\multirow{3}{*}{H20}
& BoN         & 65.5 & 34.5 & 1.41 & 36.9 & 23.1 & \textbf{1.35} & 7.1 & 1.4 & 0.63 & 29.3 & 15.9 & 0.99 \\
& GEAK        & 68.9 & 34.5 & 1.29 & 60.0 & 29.2 & 1.16 & 31.4 & 14.3 & 0.90 & 49.4 & 23.8 & 1.06 \\
& \toolname{} & \textbf{93.1} & \textbf{51.7} & \textbf{1.70} & \textbf{86.2} & \textbf{63.1} & \textbf{1.35} & \textbf{67.1} & \textbf{54.3} & \textbf{1.46} & \textbf{79.3} & \textbf{57.3} & \textbf{1.45} \\
\midrule
\multirow{3}{*}{A100}
& BoN         & 73.9 & 34.8 & 1.43 & 38.7 & 24.2 & 1.23 & 14.8 & 6.8 & 0.76 & 31.2 & 16.8 & 0.98 \\
& GEAK        & 65.2 & 43.5 & 1.09 & 54.8 & 45.2 & 1.18 & 38.6 & 33.0 & 1.54 & 48.0 & 38.7 & 1.34 \\
& \toolname{} & \textbf{95.7} & \textbf{82.6} & \textbf{2.23} & \textbf{91.9} & \textbf{77.4} & \textbf{2.05} & \textbf{67.1} & \textbf{42.1} & \textbf{1.75} & \textbf{79.8} & \textbf{60.1} & \textbf{1.91} \\
\bottomrule
\end{tabular}
\end{table*}

\paragraph{Consistent performance dominance.}
As shown in Table~\ref{tab:main_results}, \toolname{} achieves the highest performance across all GPU architectures. On A100, \toolname{} attains 1.91$\times$ geometric mean speedup with 79.8\% correctness, outperforming the best performing baseline, GEAK, by 42.5\% in speedup and 66.2\% in success rate. 
Consistent advantages can also be observed on RTX 4090 (1.74$\times$ vs. 1.44$\times$) and H20 (1.45$\times$ vs. 1.06$\times$).

\paragraph{Robust adaptation via hardware-aware optimization.}
The most effective kernel optimizations are \emph{hardware-dependent}: the optimal schedule/transform set must reflect each platform's compute-memory balance and architectural constraints~\citep{williams2009roofline,chen2018learning,zheng2020ansor}. 
Interestingly, \toolname{} does adapt its optimization choices across devices rather than applying a fixed, hardware-agnostic search policy.
For example, compared to H20, \toolname{} allocates more exploration budget to \textsc{fusion} on RTX 4090 (18.5\% vs.\ 12.8\%), while H20 explores \textsc{tiling} more frequently (10.0\% vs.\ 7.6\%). Detailed statistics and analysis can be found in Appendix~\ref{app:hardware_diversity}.
Such platform-specific optimization divergence confirms that our hardware-informed pruning (Assumption~\ref{asm:boundedness}) effectively calibrates exploration to specific bottlenecks without manual tuning.
Consequently, while baseline methods struggle with the diversity (GEAK drops from 1.44$\times$ on RTX 4090 to 1.06$\times$ on H20), \toolname{} still maintains robust performance (1.45$\times$ to 1.91$\times$) across all platforms.

\paragraph{Scaling with kernel complexity.}
The performance gap widens on challenging kernels. For the 23 \textbf{Hard (L4-5) }kernels on A100, \toolname{} achieves 67.1\% correctness and 1.75$\times$ speedup, significantly outperforming GEAK (38.6\% C, 1.54$\times$ G). While naive sampling (BoN) fails to generate valid kernels for 85\% of tasks, and unstructured search (GEAK) struggles to improve performance, \toolname{}'s structured exploration efficiently identifies paths that are both \textit{correct} and \textit{efficient} in the vast search space.

\paragraph{Practical optimization yield.}
The Fast@1 metric measures the probability of finding an \textit{optimized} kernel (speedup $>1.0\times$). 
\toolname{} demonstrates remarkable consistency, achieving Fast@1 rates of 43.3\%--60.1\% across all platforms. 
In contrast, BoN fails to accelerate most tasks (10.0\%--16.8\%), while GEAK shows both lower average performance and higher volatility (23.8\%--38.7\%). 
The 39--140\% relative improvement of \toolname{} over GEAK confirms the effectiveness of our structured exploration: \textbf{hardware-informed pruning} filters out physically invalid candidates early, and \textbf{dynamic clustering} efficiently guides the search toward high-potential regions. 
The two algorithmic designs ensure \toolname{} not only generates correct code but delivers actual speedups for a substantially larger fraction of the optimization budget.

\subsection{Detailed Analysis}
\label{sec:analysis}
In this section, We conduct a detailed analysis to validate \toolname{}'s algorithmic properties, generalization capabilities, and practical cost-efficiency. Unless otherwise stated, all experiments use a 50-kernel subset (preserving category and difficulty distribution as detailed in Appendix~\ref{app:subset}) on the H20 platform.

\subsubsection{Scalability and Clustering Sensitivity}
\label{sec:scaling}

\begin{figure}[t]
    \centering
    \includegraphics[width=0.78\linewidth]{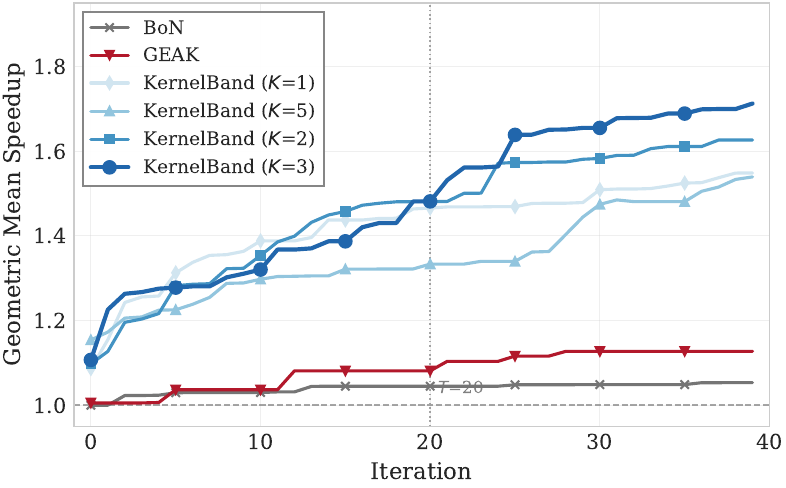}
    \caption{Scaling and clustering sensitivity. \toolname{} ($K$=3) shows sustained improvement, while baselines saturate early.}
    \label{fig:scaling}
\end{figure}

We analyze the sample efficiency (i.e., scaling to more iterations) and hyperparameter sensitivity by running \toolname{} with $K \in \{1, 2, 3, 5\}$ against baselines for an extended budget of $T=40$ iterations, as shown in Figure~\ref{fig:scaling}.

\paragraph{Superior scaling behavior.}
While baselines saturate early—BoN stagnates at 1.05$\times$ and GEAK plateaus at 1.13$\times$ after 25 iterations—\toolname{} ($K$=3) demonstrates continuous improvement, reaching 1.71$\times$ at $T$=39. At the standard budget ($T$=20), \toolname{} achieves 1.48$\times$ speedup, outperforming GEAK by 37\%. This confirms that our hierarchical MAB solution successfully allows provably efficient exploration for the challenging kernel optimization problem.

\paragraph{Optimal clustering granularity.}
The interplay between cluster count $K$ and iteration budget reveals a clear trade-off. For limited budgets ($T \le 10$), smaller $K \in \{1, 2\}$ performs best by concentrating exploration. However, as the budget grows ($T \ge 20$), $K=3$ overtakes simpler configurations (reaching 1.66$\times$ vs.\ 1.58$\times$ for $K=2$ at $T=30$). The \textbf{5\% advantage} demonstrates the value of cross-cluster knowledge transfer when sufficient budget is available. Notably, $K=5$ consistently underperforms (1.54$\times$), suggesting that excessive fragmentation hurts bandit learning. We therefore recommend $K=3$ for default optimization. Importantly, all tested $K$ configurations consistently outperform both baselines across the full iteration range, indicating that \toolname{}'s advantage is robust to this hyperparameter.

\begin{table}[t]
\centering
\caption{LLM generalization results. C: Correct (\%). G: Geometric mean speedup. F: Fast@1 (\%). Best results per model in \textbf{bold}.}
\label{tab:model_gen}
\small
\setlength{\tabcolsep}{4pt}
\begin{tabular}{llccc}
\toprule
\textbf{Model} & \textbf{Method} & \textbf{C (\%)} & \textbf{F (\%)} & \textbf{G} \\
\midrule
\multirow{3}{*}{DeepSeek-V3.2}
& BoN         & 27.5 & 12.5 & 1.10 \\
& GEAK        & 37.5 & 17.5 & 0.95 \\
& \toolname{} & \textbf{85.0} & \textbf{67.5} & \textbf{1.52} \\
\midrule
\multirow{3}{*}{GPT-5}
& BoN         & 44.9 & 28.6 & 1.14 \\
& GEAK        & 51.0 & 24.5 & 1.07 \\
& \toolname{} & \textbf{81.6} & \textbf{65.3} & \textbf{1.72} \\
\midrule
\multirow{3}{*}{Claude Opus 4.5}
& BoN         & 40.8 & 24.5 & 1.17 \\
& GEAK        & 63.3 & 38.8 & 1.30 \\
& \toolname{} & \textbf{89.8} & \textbf{73.5} & \textbf{1.82} \\
\midrule
\multirow{3}{*}{Gemini 3 Flash}
& BoN         & 47.9 & 25.0 & 1.20 \\
& GEAK        & 62.5 & 37.5 & 1.21 \\
& \toolname{} & \textbf{70.8} & \textbf{45.8} & \textbf{1.48} \\
\bottomrule
\end{tabular}
\end{table}

\subsubsection{Robustness Across LLM Backends}
\label{sec:model_gen}

To ensure our gains are not model-specific, we benchmark \toolname{} across four SOTA LLMs.

\paragraph{Generalized performance advantage.}
As shown in Table~\ref{tab:model_gen}, \toolname{} consistently outperforms baselines regardless of the underlying model. With Claude Opus 4.5, it achieves an impressive 1.82$\times$ speedup and 89.8\% correctness. Even with the smaller Gemini 3 Flash, \toolname{} maintains a substantial lead (1.48$\times$ vs. 1.21$\times$ for GEAK).

\paragraph{Compensating for Model Capabilities.}
While absolute performance naturally correlates with model strength (Claude > GPT-5 > DeepSeek > Gemini), the \textit{relative} gain provided by \toolname{} remains robust, indicating that our structured exploration framework acts as a powerful amplifier of any code LLM.

\subsection{Analysis on the strategy selection pattern.}
\paragraph{Strategy risk-reward profiles.}
Table~\ref{tab:strategy_util} reveals how the bandit policy manages trade-offs between risks and rewards. \textbf{Tiling} represents a ``high-risk, high-reward'' strategy: low success rate (14.4\%) but massive impact when successful (61.5\% best-kernel contribution). Conversely, \textbf{Vectorization} offers ``low-risk, low-reward'' gains (57.1\% success, 17.1\% best). 
\textbf{Fusion} strikes a balance (75\% success, 55\% best). The bandit policy of \toolname{} effectively navigates these profiles, prioritizing reliable gains while occasionally gambling on high-variance strategies.

\begin{table}[t]
\centering
\caption{Strategy selection statistics. Freq: selection frequency (\%). Succ: success rate (\%, correct \& speedup $>$1×). Best: percentage of successful applications that contributed to the final best kernel.}
\label{tab:strategy_util}
\small
\setlength{\tabcolsep}{4pt}
\begin{tabular}{lccc}
\toprule
\textbf{Strategy} & \textbf{Freq (\%)} & \textbf{Succ (\%)} & \textbf{Best (\%)} \\
\midrule
Tiling & 10.0 & 14.4 & 61.5 \\
Vectorization & 14.7 & 57.1 & 17.1 \\
Fusion & 12.8 & 75.0 & 55.2 \\
Pipeline & 9.8 & 64.4 & 26.3 \\
Reordering & 33.2 & 48.7 & 25.3 \\
Access \& Layout & 19.5 & 29.5 & 19.2 \\
\bottomrule
\end{tabular}
\end{table}

\subsubsection{Cost and Efficiency Analysis}
\label{sec:cost}

Finally, we examine the computational and economic feasibility of \toolname{} for real-world use.

\begin{figure}[t]
    \centering
    \includegraphics[width=0.8\linewidth]{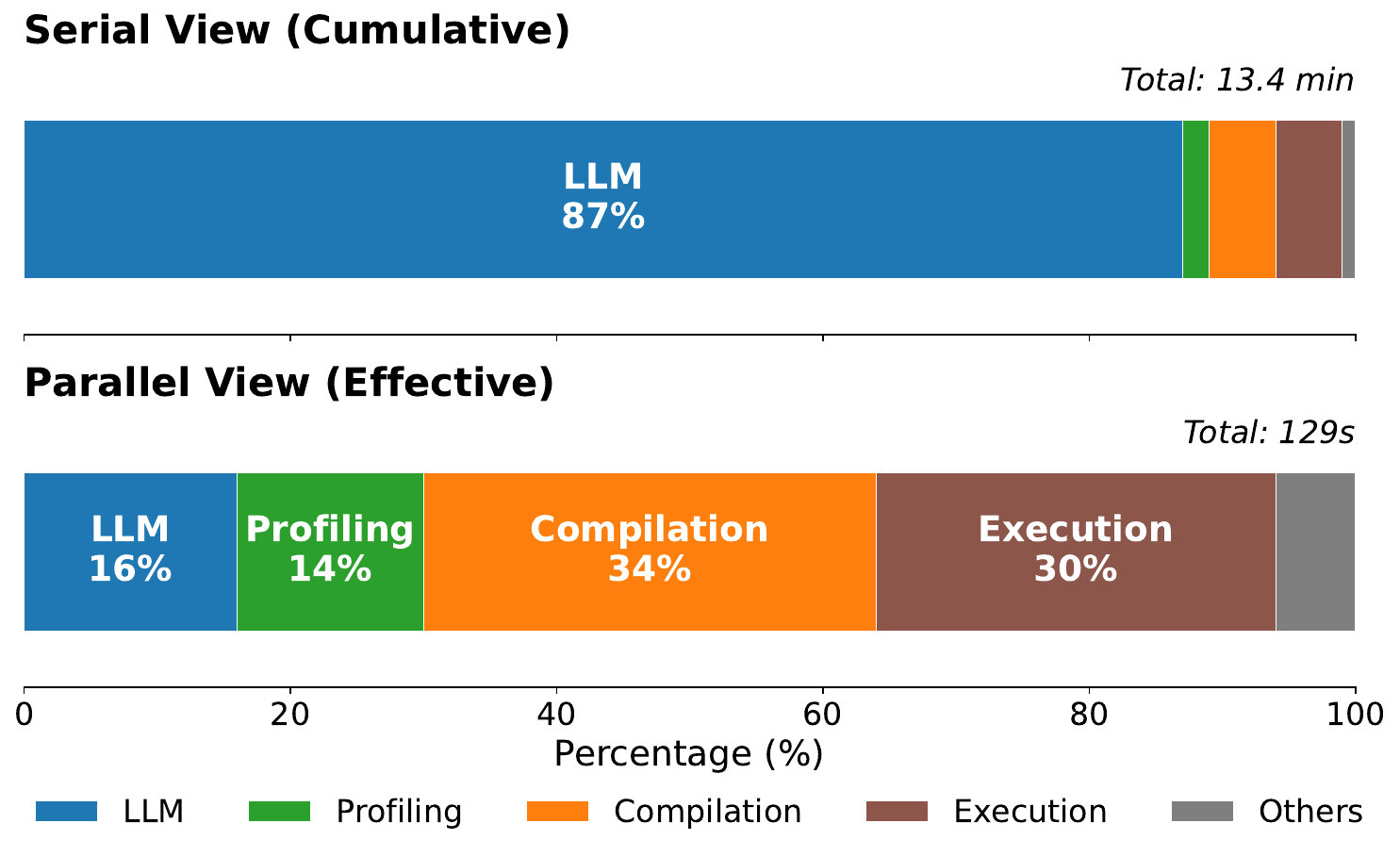}
    \caption{Time breakdown per kernel/iteration: (a) Serial cumulative time (13.4 min); (b) Parallel wall-clock time with batched LLM calls (129s).}
    \label{fig:time_breakdown}
\end{figure}

\paragraph{Time breakdown.}
Figure~\ref{fig:time_breakdown} decomposes the per-kernel iteration latency. While LLM inference dominates serial execution (87\%), batched inference shifts the bottleneck to compilation (34\%) and execution (30\%), reducing the effective wall-clock time to just 129s per iteration. This confirms that \toolname{} fits comfortably within practical compilation timeouts.

\begin{figure}[t]
    \centering
    \includegraphics[width=0.75\linewidth]{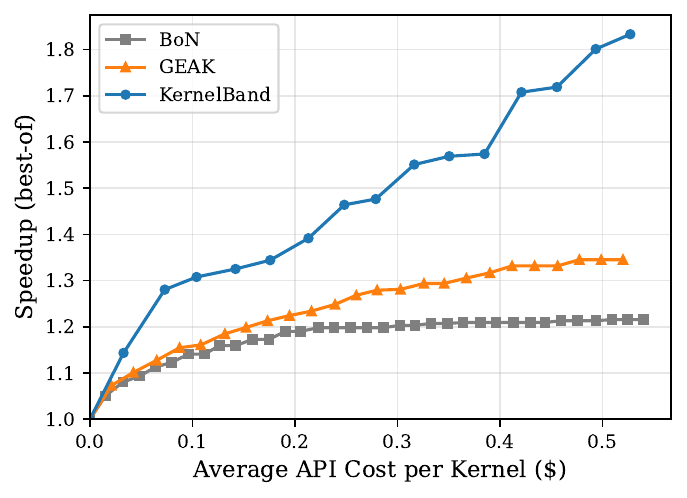}
    \caption{Speedup vs.\ API cost per kernel. \toolname{} yields 35-50\% higher speedup at equivalent budgets.}
    \label{fig:api_cost}
\end{figure}

\paragraph{Cost analysis.}
Despite higher per-iteration costs due to multi-strategy exploration, \toolname{} delivers superior optimization per dollar as shown in Figure~\ref{fig:api_cost}. At a fixed budget of \$0.50 per kernel, it achieves 1.83$\times$ speedup, outperforming GEAK by 35\% (1.35$\times$) and BoN by 50\% (1.22$\times$). 
This cost-effectiveness stems from our pruning mechanism, filtering low-value candidates early, ensuring that API costs translate directly to performance gains.

\subsection{Ablation Study}
\label{sec:ablation}

We dissect the effectiveness of \toolname{} through single-component ablations (removing one module) and framework-level changes (altering the optimization paradigm).
Detailed descriptions and analysis of ablation settings can be found in Appendix~\ref{app:ablation_detail_setting}.

\paragraph{Structured bandit policy is foundational.}
As shown in Table~\ref{tab:ablation}, the most critical finding is that replacing bandit-based selection with LLM semantic reasoning (\textit{LLM Strategy Selection}, where a LLM selects which strategy to apply based on kernel analysis rather than using a bandit policy) causes a catastrophic drop to 0.97$\times$ speedup. 
This confirms that learned bandit policy are superior to LLM intuition for strategy selection. Furthermore, removing the strategy set entirely (\textit{w/o Strategy}, where free-form iterative generation is used as GEAK without structured strategies or profiling guidance) drops performance to 1.15$\times$. 
Injecting raw profiling metrics without strategies (\textit{+ Raw Profiling}) introduces noise, further degrading correctness to 43.9\%. These results validate that our structured bandit policy is the essential bridge between hardware information and code generation.

\paragraph{Component contribution hierarchy.}
As shown in Table~\ref{tab:ablation}, at the standard budget ($T=20$), profiling guidance proves more critical than clustering: disabling profiling (\textit{w/o Profiling}) causes a 20\% speedup drop (1.57$\times \to$ 1.26$\times$), while disabling clustering (\textit{w/o Clustering}) yields a 10\% drop (1.41$\times$). However, as shown in Section~\ref{sec:scaling}, clustering's value grows with iteration budget. Both components contribute incrementally to the 2.6$\times$ improvement over the BoN baseline (0.60$\times$).

\begin{table}[t]
\centering
\caption{Single-component ablations show minimal drops; framework-level ablations show the value of core designs.}
\label{tab:ablation}
\small
\setlength{\tabcolsep}{4pt}
\begin{tabular}{llccc}
\toprule
\textbf{Type} & \textbf{Configuration} & \textbf{C (\%)} & \textbf{F (\%)} & \textbf{G} \\
\midrule
\multirow{4}{*}{\rotatebox[origin=c]{90}{\scriptsize Single}}
& KernelBand (Full) & 87.8 & 63.4 & 1.57 \\
& \quad w/o Clustering ($K$=1) & 82.9 & 58.5 & 1.41 \\
& \quad w/o Profiling & 85.4 & 56.1 & 1.26 \\
& \quad LLM Strategy Selection & 68.3 & 36.6 & 0.97 \\
\midrule
\multirow{3}{*}{\rotatebox[origin=c]{90}{\scriptsize Frame.}}
& w/o Strategy + Raw Prof. & 43.9 & 26.8 & 1.12 \\
& w/o Strategy Set & 78.0 & 48.8 & 1.15 \\
& BoN (baseline) & 34.2 & 17.1 & 0.60 \\
\bottomrule
\end{tabular}
\end{table}

%% file: related_work.tex
\section{Related Work}

\textbf{LLM-based kernel optimization.}
Several recent and concurrent works explore LLM-based agentic workflows for automated kernel optimization.
STARK~\citep{dong2025starks} employs multi-agent collaboration with grounded instruction and strategic search to explore the optimization space. CudaForge~\citep{cudaforge} uses a Coder-Judge architecture that iteratively generates and refines CUDA kernels with Nsight Compute profiling feedback. GEAK~\citep{wang2025geak} adapts Reflexion-style reasoning loops with inference-time compute scaling for Triton kernel generation. TritonForge~\citep{tritonforge} integrates runtime profiling with iterative code transformation to identify bottlenecks and propose targeted modifications. 
While these methods demonstrate the value of code LLMs, they lack principled exploration-exploitation mechanisms to effectively navigate the vast optimization space of kernels.

An alternative paradigm focuses on fine-tuning LLMs specifically for kernel generation. ConCuR~\citep{kong2025concur} develops a pipeline to generate and curate CUDA kernels with reasoning traces for supervised fine-tuning. Kevin~\citep{baronio2025kevin} applies multi-turn RL to address challenges in long-trajectory kernel optimization. TritonRL~\citep{woo2025tritonrl} combines supervised fine-tuning with RL using hierarchical reward assignment to mitigate reward hacking. Our work is orthogonal to and complement this line of work, as \toolname{} serves as an amplifier of any given code LLMs demonstrated in our experiment.

\textbf{Multi-armed bandits and their applications.} Multi-armed bandits (MABs) \citep{berry1997bandit,slivkins2019introduction,kuleshov2014algorithms,vermorel2005multi} provide a principled framework for sequential decision-making under uncertainty, with applications ranging from clinical trials to online advertising~\citep{lai1985asymptotically}. Classical algorithms like UCB~\citep{auer2002finite} and Thompson Sampling~\citep{thompson1933likelihood} offer optimal regret bounds for stationary environments, while recent work has extended these to contextual~\citep{li2010contextual,chu2011contextual}, metric~\citep{kleinberg2008multi}, clustering~\citep{gentile2014online}, and hierarchical settings~\citep{bubeck2011xarmed}. 
In systems optimization, bandit algorithms have shown promise for parameter tuning~\citep{snoek2012practical,pacula2012hyperparameter}, compiler~\citep{xu2017parallel}, and resource allocation~\citep{pandey2025multi}. However, applying bandit algorithms to kernel optimization is largely unexplored, despite the natural fit between the exploration-exploitation trade-off and the challenge of navigating vast optimization spaces.

%% file: conclusion.tex
\section{Conclusion}
We have introduced \toolname{}, a framework that bridges the gap between code LLMs and kernel optimization with a hardware-constrained contextual bandit formulation. 
By combining runtime behavior clustering with profiling-guided pruning, \toolname{} enables provably efficient navigation of the vast optimization space. 
Extensive experiments demonstrate consistent improvements over state-of-the-art methods, achieving up to 1.91$\times$ speedup with 39--140\% higher optimization Fast@1 across diverse hardware platforms and LLM backends. Our work highlights the importance of structured and hardware-aware exploration to unlocking LLMs' potential for system optimization.

%% file: appendix.tex
%%%%%%%%%%%%%%%%%%%%%%%%%%%%%%%%%%%%%%%%%%%%%%%%%%%%%%%%%%%%%%%%%%%%%%%%%%%%%%%
%%%%%%%%%%%%%%%%%%%%%%%%%%%%%%%%%%%%%%%%%%%%%%%%%%%%%%%%%%%%%%%%%%%%%%%%%%%%%%%
% APPENDIX
%%%%%%%%%%%%%%%%%%%%%%%%%%%%%%%%%%%%%%%%%%%%%%%%%%%%%%%%%%%%%%%%%%%%%%%%%%%%%%%
%%%%%%%%%%%%%%%%%%%%%%%%%%%%%%%%%%%%%%%%%%%%%%%%%%%%%%%%%%%%%%%%%%%%%%%%%%%%%%%
\newpage
\appendix
\onecolumn

\section{Implementation Details}
\label{app:impl}

\subsection{Profiling and Clustering Features}
\label{app:profiling_metrics}
\label{app:clustering_features}

\paragraph{Behavioral feature vector $\phi(k)$.} As defined in Section~\ref{subsec:characterization}, we extract a 5-dimensional vector for clustering kernels with similar optimization responses:
\begin{itemize}[leftmargin=1.5em, itemsep=0pt]
    \item $\tilde{\mathcal{T}}(k)$: normalized execution time (from timing run, log-transformed)
    \item $n_{\text{reg}}$: registers per thread (from \texttt{cuFuncGetAttribute})
    \item $n_{\text{smem}}$: shared memory per block (from \texttt{cuFuncGetAttribute})
    \item $d_{\text{block}}$: block dimensions (from launch configuration)
    \item $\eta_{\text{occ}}$: theoretical occupancy (computed via \texttt{cudaOccupancyMaxActiveBlocksPerMultiprocessor})
\end{itemize}

\paragraph{Hardware signature $h(k)$.} For hardware-aware pruning, we extract throughput metrics via NVIDIA Nsight Compute to identify resource saturation:
\begin{itemize}[leftmargin=1.5em, itemsep=0pt]
    \item SM throughput (\texttt{sm\_\_throughput.avg.pct\_of\_peak\_sustained\_elapsed})
    \item DRAM throughput (\texttt{dram\_\_throughput.avg.pct\_of\_peak\_sustained\_elapsed})
    \item L2 throughput (\texttt{lts\_\_throughput.avg.pct\_of\_peak\_sustained\_elapsed})
\end{itemize}
These three metrics correspond to the compute, memory bandwidth, and cache bottleneck categories described in Section~\ref{subsec:characterization}.

\section{Proof of Theorem~\ref{thm:convergence}}
\label{app:theory}

In this section, we provide the detailed derivation for the convergence bound of \toolname{}.

\subsection{Regret Decomposition}
Let $x^* = \arg\max_{x \in \mathcal{X}} \mu(x)$ be the global optimal kernel configuration with expected reward $\mu^*$.
Let $x^*_{disc}$ be the best candidate available in our discretized cluster centers, i.e., $x^*_{disc} = \arg\max_{c \in \{C_1, \dots, C_K\}} \mu(c)$.

The cumulative regret $R(T)$ after $T$ rounds can be decomposed into \textit{Approximation Regret} ($R_{approx}$) and \textit{Estimation Regret} ($R_{est}$):

\begin{align}
R(T) &= \sum_{t=1}^T \left( \mu^* - \mu(x_t) \right) \\
     &= \sum_{t=1}^T \left( \mu^* - \mu(x^*_{disc}) + \mu(x^*_{disc}) - \mu(x_t) \right) \\
     &= \underbrace{T \cdot (\mu^* - \mu(x^*_{disc}))}_{R_{approx}} + \underbrace{\sum_{t=1}^T (\mu(x^*_{disc}) - \mu(x_t))}_{R_{est}}
\end{align}

\subsection{Bounding Approximation Regret}
Based on Assumption~\ref{asm:smoothness} (Lipschitz Continuity), the performance function $\mu(\cdot)$ is $L$-Lipschitz. Since our clustering algorithm covers the space such that for any $x$, there exists a cluster center $c$ with $\|x - c\| \le \epsilon$, the gap between the global optimum and the best discrete representative is bounded by:
\[
\mu^* - \mu(x^*_{disc}) \le L \cdot \|x^* - x^*_{disc}\| \le L\epsilon
\]
Therefore:
\begin{equation}
R_{approx} \le T \cdot L\epsilon
\end{equation}

\subsection{Bounding Estimation Regret}
The term $R_{est}$ represents the regret incurred by a Multi-Armed Bandit (MAB) algorithm selecting among $K$ clusters (and their associated strategies $|\mathcal{S}|$). The total number of arms is $N = K|\mathcal{S}|$.
For standard UCB algorithms, the regret is bounded by $\tilde{O}(\sqrt{N T})$. Specifically, with probability $1-\delta$:
\begin{equation}
R_{est} \le \sqrt{C \cdot K|\mathcal{S}| T \log(T/\delta)}
\end{equation}
for some universal constant $C$.

\subsection{Final Average Regret Bound}
Combining the bounds for $R_{approx}$ and $R_{est}$:
\[
R(T) \le O\left(\sqrt{K|\mathcal{S}| T \log T}\right) + L\epsilon T
\]
Dividing by $T$ to obtain the average regret (convergence rate):
\[
\frac{R(T)}{T} \le O\left(\sqrt{\frac{K|\mathcal{S}| \log T}{T}}\right) + L\epsilon
\]
As $T \to \infty$, the first term approaches 0, ensuring that the algorithm converges to the $L\epsilon$-neighborhood of the global optimum.

\section{LLM Configuration}
\label{app:llm_config}

Table~\ref{tab:llm_config} summarizes the configuration for all LLM backends used in our experiments.

\begin{table}[h]
\centering
\caption{LLM backend configurations.}
\label{tab:llm_config}
\small
\begin{tabular}{ll}
\toprule
\textbf{Parameter} & \textbf{Value} \\
\midrule
Primary Model & DeepSeek-V3.2 \\
Access Method & Official API \\
Temperature & 1.0 \\
Max Output Tokens & 16384 \\
\bottomrule
\end{tabular}
\end{table}

\section{Optimization Strategy Set}
\label{app:strategies}

Table~\ref{tab:strategies_full} provides the complete list of optimization strategies in $\mathcal{S}$ with their descriptions and example transformations.

\begin{table}[h]
\centering
\caption{Complete optimization strategy set. This refined set was distilled from an initial 10 strategies through pilot experiments.}
\label{tab:strategies_full}
\small
\begin{tabular}{lp{5.5cm}}
\toprule
\textbf{Strategy} & \textbf{Description} \\
\midrule
Tiling & Partition computation into configurable tile sizes for improved cache locality and parallelism \\
Vectorization & Use vector loads/stores (e.g., float4) for improved memory throughput \\
Fusion & Combine multiple operations to reduce intermediate memory traffic \\
Pipeline & Configure software pipelining depth for latency hiding \\
Reordering & Optimize loop order and instruction scheduling for better ILP \\
Access \& Layout & Optimize memory access patterns, coalescing, and data layout \\
\bottomrule
\end{tabular}
\end{table}

\section{Benchmark Subset Details}
\label{app:subset}

For detailed analysis experiments, we use a stratified subset of 50 kernels from the 184-kernel TritonBench-G benchmark. The subset was generated using stratified random sampling (seed=42) to preserve the original category and difficulty distribution:

\begin{itemize}
    \item \textbf{Category coverage}: All 13 functional categories represented
    \item \textbf{Difficulty distribution}: L1 (1), L2 (7), L3 (18), L4 (23), L5 (1)
    \item \textbf{Sampling ratio}: 27.2\% of the full benchmark
    \item \textbf{Maximum deviation from original distribution}: $<$1\%
\end{itemize}

Table~\ref{tab:subset_distribution} compares the category distribution between the full benchmark and the sampled subset. Table~\ref{tab:subset_kernels} lists all 50 kernels organized by difficulty level and functional category.

\begin{table}[t]
\centering
\caption{Category distribution: full benchmark vs. sampled subset.}
\label{tab:subset_distribution}
\small
\begin{tabular}{lcc}
\toprule
\textbf{Category} & \textbf{Full (184)} & \textbf{Subset (50)} \\
\midrule
Attention & 29 (15.8\%) & 7 (14.0\%) \\
MatMul/GEMM & 26 (14.1\%) & 7 (14.0\%) \\
Normalization & 18 (9.8\%) & 4 (8.0\%) \\
Linear Attention/SSM & 17 (9.2\%) & 4 (8.0\%) \\
Element-wise Ops & 16 (8.7\%) & 3 (6.0\%) \\
Memory/Index Ops & 13 (7.1\%) & 3 (6.0\%) \\
Other & 12 (6.5\%) & 3 (6.0\%) \\
Embedding/RoPE & 11 (6.0\%) & 3 (6.0\%) \\
Softmax & 11 (6.0\%) & 4 (8.0\%) \\
Fused Ops/Activation & 10 (5.4\%) & 4 (8.0\%) \\
Quantization & 8 (4.3\%) & 2 (4.0\%) \\
Loss Functions & 7 (3.8\%) & 3 (6.0\%) \\
Reduction & 6 (3.3\%) & 3 (6.0\%) \\
\bottomrule
\end{tabular}
\end{table}

\begin{table}[t]
\centering
\caption{Complete list of 50 kernels in the evaluation subset, stratified by difficulty (L1--L5) and functional category.}
\label{tab:subset_kernels}
\small
\begin{tabular}{clll}
\toprule
\textbf{\#} & \textbf{Diff.} & \textbf{Category} & \textbf{Kernel Name} \\
\midrule
1 & L1 & Element-wise Ops & cosine\_compute \\
\midrule
2 & L2 & Attention & flash\_decode2\_phi \\
3 & L2 & MatMul/GEMM & matmul\_kernel \\
4 & L2 & Memory/Index Ops & matrix\_transpose \\
5 & L2 & Normalization & triton\_mul2 \\
6 & L2 & Other & square\_matrix \\
7 & L2 & Reduction & triton\_argmax \\
8 & L2 & Softmax & softmax\_triton1 \\
\midrule
9 & L3 & Attention & flash\_decode2\_llama \\
10 & L3 & Element-wise Ops & pow\_scalar\_tensor \\
11 & L3 & Embedding/RoPE & embedding\_triton\_kernel \\
12 & L3 & Fused Ops/Activation & relu\_strided\_buffer \\
13 & L3 & Fused Ops/Activation & swiglu\_backward \\
14 & L3 & Fused Ops/Activation & swiglu\_triton \\
15 & L3 & Linear Attention/SSM & chunk\_cumsum\_vector \\
16 & L3 & Linear Attention/SSM & reversed\_cumsum\_scalar \\
17 & L3 & Loss Functions & kldiv\_triton \\
18 & L3 & MatMul/GEMM & triton\_matmul \\
19 & L3 & Memory/Index Ops & var\_len\_copy \\
20 & L3 & Normalization & layer\_norm\_welfold \\
21 & L3 & Normalization & rmsnorm\_fused\_llama \\
22 & L3 & Other & uniform\_sampling \\
23 & L3 & Quantization & quantize\_kv\_copy \\
24 & L3 & Reduction & matrix\_reduction \\
25 & L3 & Softmax & softmax\_triton2 \\
26 & L3 & Softmax & softmax\_triton3 \\
\midrule
27 & L4 & Attention & attention\_fwd\_triton1 \\
28 & L4 & Attention & attention\_fwd\_triton2 \\
29 & L4 & Attention & attention\_kernel \\
30 & L4 & Attention & triton\_attention \\
31 & L4 & Element-wise Ops & matrix\_vector\_multip \\
32 & L4 & Embedding/RoPE & fast\_rope\_embedding \\
33 & L4 & Embedding/RoPE & rope\_backward\_transform \\
34 & L4 & Fused Ops/Activation & relu\_triton\_kernel \\
35 & L4 & Linear Attention/SSM & chunk\_gate\_recurrence \\
36 & L4 & Linear Attention/SSM & fused\_recurrent\_retention \\
37 & L4 & Loss Functions & cross\_entropy\_ops \\
38 & L4 & Loss Functions & fast\_ce\_loss \\
39 & L4 & MatMul/GEMM & int8\_matmul\_quantization \\
40 & L4 & MatMul/GEMM & int\_scaled\_matmul \\
41 & L4 & MatMul/GEMM & matmul\_dequantize\_int4 \\
42 & L4 & MatMul/GEMM & rms\_matmul\_rbe \\
43 & L4 & MatMul/GEMM & streamk\_matmul \\
44 & L4 & Memory/Index Ops & kcache\_copy\_triton \\
45 & L4 & Normalization & fused\_layernorm\_triton \\
46 & L4 & Other & bgmv\_expand\_slice \\
47 & L4 & Quantization & quantize\_copy\_kv \\
48 & L4 & Reduction & logsumexp\_fwd \\
49 & L4 & Softmax & ksoftmax\_triton \\
\midrule
50 & L5 & Attention & context\_attn\_bloom \\
\bottomrule
\end{tabular}
\end{table}

\section{Baseline Discussion}
\label{app:baseline_discussion}

\textbf{Benchmark Modifications.}
The original TritonBench release contained implementation issues that impacted reliable evaluation.
We adopt the corrected benchmark version provided by GEAK~\citep{wang2025geak}, which addresses these issues.
Additionally, we performed straightforward function substitutions to convert AMD-specific implementations to their NVIDIA equivalents, without modifying kernel logic or semantics.

\textbf{GEAK Adaptation.}
GEAK's adaptation to NVIDIA GPUs involved only platform-specific configurations (e.g., device queries, memory bandwidth specifications) without altering the agent's core logic or prompts.

\textbf{Other Agent-based Methods.}
Other agent-based methods present reproducibility challenges: STARK~\citep{dong2025starks}, TritonForge, and QiMeng-Kernel have not released their code; we contacted authors but code remains unavailable due to corporate IP constraints or ongoing submissions. CudaForge~\citep{cudaforge} targets CUDA kernels on KernelBench and would require non-trivial adaptation to our Triton-based evaluation framework; we consider it concurrent work.

\section{PyTorch Baseline Comparison}
\label{app:pytorch_baseline}

To contextualize the optimization gains of Triton kernels relative to standard PyTorch execution, we compare \toolname{}-optimized kernels against three PyTorch execution modes: (1) \textbf{eager} mode (default PyTorch execution), (2) \textbf{inductor} (\texttt{torch.compile} with the default inductor backend), and (3) \textbf{max-autotune} (\texttt{torch.compile} with \texttt{mode="max-autotune"}).

\paragraph{Experimental setup.}
From the 50-kernel subset, we select 30 kernels suitable for fair comparison with PyTorch, where native operators are available (e.g., \texttt{torch.softmax}, \texttt{F.layer\_norm}). We exclude special-purpose kernels (e.g., Flash Attention decode, INT4 quantization, LoRA operations) that lack general PyTorch counterparts. Experiments are conducted on H20 with DeepSeek-V3.2 and $T=20$ iterations.

\paragraph{Results.}
Table~\ref{tab:pytorch_baseline} presents the geometric mean speedup of \toolname{}-optimized Triton kernels over each PyTorch baseline. The results demonstrate that \toolname{} achieves substantial speedups over all PyTorch execution modes, with 1.87$\times$ over inductor, 2.16$\times$ over max-autotune, and 2.13$\times$ over eager execution. The larger speedup over max-autotune compared to inductor is noteworthy: while max-autotune performs more extensive autotuning, it may over-specialize to specific input shapes, whereas \toolname{}'s optimization generalizes better across the diverse input shapes in our evaluation.

\begin{table}[h]
\centering
\caption{Speedup of \toolname{}-optimized Triton kernels over PyTorch baselines (30 kernels, H20, $T=20$).}
\label{tab:pytorch_baseline}
\small
\begin{tabular}{lc}
\toprule
\textbf{PyTorch Baseline} & \textbf{Speedup} \\
\midrule
vs.\ eager & 2.13$\times$ \\
vs.\ inductor & 1.87$\times$ \\
vs.\ max-autotune & 2.16$\times$ \\
\bottomrule
\end{tabular}
\end{table}

These results validate that \toolname{}'s LLM-driven Triton kernel optimization provides meaningful performance gains beyond what PyTorch's built-in compilation and autotuning can achieve, justifying the investment in custom kernel development for performance-critical workloads.

\section{Evaluation Protocol Details}
\label{app:eval_details}

\paragraph{Correctness verification.}
The \textit{Execution Accuracy} check uses \texttt{torch.allclose} with
absolute tolerance $\text{atol}=10^{-4}$ and relative tolerance
$\text{rtol}=10^{-4}$. A kernel passes when
$|\text{generated} - \text{reference}| \leq \text{atol} + \text{rtol} \times |\text{reference}|$.

\paragraph{Statistical robustness.}
For each input shape, we use Triton's \texttt{triton.testing.do\_bench}
function, which handles common GPU benchmarking pitfalls such as
measuring only kernel launch time (via \texttt{time.time}), omitting
cache clearing, and skipping warmup. The function performs 100ms of
warmup runs to stabilize GPU state, followed by 1000ms of timed runs.
We report the median execution time to reduce sensitivity to outliers.

\paragraph{Weighted aggregation.}
The overall kernel speedup is computed as the ratio of total baseline runtime to total optimized runtime:
\[
\text{Speedup} = \frac{\sum_i t_{\text{baseline},i}}{\sum_i t_{\text{optimized},i}}
\]
This metric reflects end-to-end performance: shapes with longer execution times naturally dominate the aggregation.

\section{Robust adaptation to hardware diversity}
\label{app:hardware_diversity}

Prior work has repeatedly shown that the most effective operator optimizations are \emph{hardware dependent}: the optimal schedule/transform set must reflect each platform's compute--memory balance and microarchitectural constraints (e.g., cache behavior, memory bandwidth, and parallel execution resources) \citep{williams2009roofline,chen2018learning,zheng2020ansor}. Consistent with this principle, our empirical strategy statistics provide direct evidence that \toolname{} adapts its optimization choices across devices rather than applying a fixed, hardware-agnostic search policy. Table~\ref{tab:strategy_stats_h20_4090} reports three complementary views: (i) selection frequency (\textsc{Freq}), (ii) success rate among attempted transformations (\textsc{Succ}; correctness and speedup $>1\times$), and (iii) contribution to the final best kernel (\textsc{Best}). The observed shifts in \textsc{Freq} across H20 and RTX~4090 indicate that \toolname{} reallocates exploration budget across strategy families in a platform-aware manner, aligning with the long-standing motivation behind hardware-targeted auto-scheduling systems \citep{chen2018learning,zheng2020ansor}.

Concretely, the strategy mix differs noticeably between H20 and RTX~4090. For example, \textsc{Fusion} is selected substantially more often on RTX~4090 (\textsc{Freq} 18.5\%) than on H20 (12.8\%), and it remains highly reliable on both devices (\textsc{Succ} 78.6\% on RTX~4090 vs.\ 75.0\% on H20), while also contributing frequently to the best-performing kernels (\textsc{Best} 63.1\% vs.\ 55.2\%). In contrast, \textsc{Tiling} exhibits an inverse risk--reward profile: it is attempted slightly more on H20 (10\%) than on RTX~4090 (7.6\%), yet has low success rates on both (14.4\% vs.\ 18.7\%) while yielding disproportionately large best-kernel contributions when it does succeed (\textsc{Best} 61.5\% on H20 vs.\ 47.3\% on RTX~4090). Meanwhile, \textsc{Access \& Layout} becomes more prominent on RTX~4090 (22.9\% vs.\ 19.5\%) and shows higher effectiveness there (\textsc{Succ} 38.9\% vs.\ 29.5\%), suggesting device-specific sensitivity to memory access behavior and layout-driven locality. These cross-device changes collectively support our claim that \toolname{} calibrates its search to the dominant bottlenecks of each platform, as predicted by performance modeling insights (e.g., Roofline) and confirmed by hardware-aware optimization studies \citep{williams2009roofline,tschand2025swizzleperf}.

\begin{table*}[t]
  \centering
  \caption{Strategy utilization statistics across H20 and RTX~4090 (50 kernels, $T{=}20$). \textsc{Freq}: selection frequency (\%). \textsc{Succ}: success rate (\%, correct and speedup $>1\times$). \textsc{Best}: percentage of successful applications that contributed to the final best kernel.}
  \label{tab:strategy_stats_h20_4090}
  \begin{subtable}[t]{0.48\textwidth}
    \centering
    \caption{H20}
    \label{tab:strategy_stats_h20}
    \begin{tabular}{lccc}
      \toprule
      \textbf{Strategy} & \textbf{Freq (\%)} & \textbf{Succ (\%)} & \textbf{Best (\%)} \\
      \midrule
      Tiling           & 10    & 14.4 & 61.5 \\
      Vectorization    & 14.7  & 57.1 & 17.1 \\
      Fusion           & 12.8  & 75.0 & 55.2 \\
      Pipeline         & 9.8   & 64.4 & 26.3 \\
      Reordering       & 33.2  & 48.7 & 25.3 \\
      Access \& Layout & 19.5  & 29.5 & 19.2 \\
      \bottomrule
    \end{tabular}
  \end{subtable}
  \hfill
  \begin{subtable}[t]{0.48\textwidth}
    \centering
    \caption{RTX 4090}
    \label{tab:strategy_stats_4090}
    \begin{tabular}{lccc}
      \toprule
      \textbf{Strategy} & \textbf{Freq (\%)} & \textbf{Succ (\%)} & \textbf{Best (\%)} \\
      \midrule
      Tiling           & 7.6   & 18.7 & 47.3 \\
      Vectorization    & 16.2  & 61.4 & 14.6 \\
      Fusion           & 18.5  & 78.6 & 63.1 \\
      Pipeline         & 5.7   & 51.9 & 13.8 \\
      Reordering       & 29.1  & 49.8 & 30.7 \\
      Access \& Layout & 22.9  & 38.9 & 35.9 \\
      \bottomrule
    \end{tabular}
  \end{subtable}
\end{table*}

\section{Detailed Ablation Study}
\label{app:ablation_detail_setting}

\subsubsection{Component Ablation}
\label{sec:component_ablation}

To understand the contribution of each component, we evaluate \toolname{} with seven configurations organized into two categories: \emph{single-component ablations} that remove one component while preserving others, and \emph{framework-level ablations} that alter the fundamental optimization paradigm.

\paragraph{Single-component ablations.}
These configurations isolate the contribution of individual components:
\begin{itemize}[itemsep=2pt]
    \item \textbf{KernelBand (Full)}: Complete system with strategy set, hardware-aware masking and kernel selection, UCB-based exploration, and runtime clustering ($K=3$).
    \item \textbf{w/o Clustering} ($K=1$): All kernels treated as a single cluster. Tests whether runtime-behavior clustering enables effective cross-kernel knowledge transfer.
    \item \textbf{w/o Profiling}: Hardware masking disabled ($M_{i,s}=1$ for all), and within-cluster kernel selection falls back to recency tie-break. Tests the value of hardware profiling guidance.
    \item \textbf{LLM Strategy Selection}: Strategy set preserved, but the LLM selects which strategy to apply based on kernel analysis rather than UCB statistics. Tests whether learned statistics outperform LLM semantic judgment.
\end{itemize}

\paragraph{Framework-level ablations.}
These configurations fundamentally alter how optimization is structured:
\begin{itemize}[itemsep=2pt]
    \item \textbf{w/o Strategy + Raw Profiling}: Removes the strategy set; the LLM generates optimizations freely with raw profiling metrics (L2 miss rate, memory bandwidth, etc.) injected into the prompt. Tests structured strategy abstraction versus raw metric injection~\citep{cudaforge}.
    \item \textbf{w/o Strategy Set}: Free-form iterative generation without structured strategies or profiling guidance, similar to Reflexion-style approaches~\citep{shinn2023reflexion}. Since the strategy set is foundational to UCB statistics and profiling compatibility, removing it effectively disables these components.
    \item \textbf{BoN (baseline)}: Best-of-N independent sampling without iteration. Included as a lower bound to quantify the value of iterative optimization.
\end{itemize}

\paragraph{Component dependencies.}
An important insight is that the strategy set $\mathcal{S}$ serves as the foundation for other components: profiling computes compatibility $\psi(s, \phi(k))$ per strategy, UCB maintains statistics $\hat{\mu}_{i,s}$ per (cluster, strategy) pair, and clustering's primary value is enabling cross-cluster UCB statistic sharing. Thus, removing the strategy set (framework-level ablation) implicitly disables structured use of profiling and UCB, which is why we expect a larger performance gap for framework-level ablations compared to single-component ablations.